\documentclass[journal]{IEEEtran}
\usepackage{lipsum} 
\usepackage{enumitem} 
\usepackage{multicol,multienum}
\usepackage{breqn}
\usepackage{stfloats}
\usepackage{multirow}
\usepackage{diagbox}
\usepackage[hidelinks]{hyperref} 
\usepackage{bm}
\usepackage{cite}
\usepackage{graphicx}
\usepackage{tabularx}
\usepackage[ruled,vlined]{algorithm2e} 
\usepackage{algorithmic}
\usepackage{booktabs}
\usepackage{amsthm}

\theoremstyle{definition}

\usepackage[tight,footnotesize]{subfigure}
\usepackage{amssymb}
\usepackage{amsmath}
\usepackage{color}
\usepackage{siunitx}
\usepackage{mathtools}

\usepackage{cleveref}
\crefname{section}{Sec.}{Sec.}
\crefname{figure}{Fig.}{Fig.}
\crefname{table}{Table}{Table}
\crefname{algorithm}{Algorithm}{Algorithm}
\crefname{equation}{Eq.}{Eq.}

\usepackage[defaultcolor=red,addedmarkup=uline]{changes}
\setcommentmarkup{\footnote{\color{authorcolor} #1}}

\usepackage{soul}
\soulregister\cite7
\soulregister\ref7
\soulregister\pageref7

\newcommand{\multirowoffset}{-0.5\dimexpr \aboverulesep + \belowrulesep + \cmidrulewidth}

\ifCLASSINFOpdf

\else

\fi

\begin{document}\sloppy


\title{Enhancing Traffic Prediction with Learnable\\ Filter Module}

\author{
Yuanshao~Zhu,
Yongchao~Ye,
Xiangyu~Zhao\textsuperscript{$\dagger$} and
James~J.Q.~Yu\textsuperscript{$\dagger$}

}

\markboth{}%
{Shell \MakeLowercase{\textit{et al.}}: Bare Demo of IEEEtran.cls for IEEE Journals}

\maketitle

\begin{abstract}
Modeling future traffic conditions often relies heavily on complex spatial-temporal neural networks to capture spatial and temporal correlations, which can overlook the inherent noise in the data. 
This noise, often manifesting as unexpected short-term peaks or drops in traffic observation, is typically caused by traffic accidents or inherent sensor vibration.
In practice, such noise can be challenging to model due to its stochastic nature and can lead to overfitting risks if a neural network is designed to learn this behavior. 
To address this issue, we propose a learnable filter module to filter out noise in traffic data adaptively. 
This module leverages the Fourier transform to convert the data to the frequency domain, where noise is filtered based on its pattern. 
The denoised data is then recovered to the time domain using the inverse Fourier transform.
Our approach focuses on enhancing the quality of the input data for traffic prediction models, which is a critical yet often overlooked aspect in the field. 
We demonstrate that the proposed module is lightweight, easy to integrate with existing models, and can significantly improve traffic prediction performance. 
Furthermore, we validate our approach with extensive experimental results on real-world datasets, showing that it effectively mitigates noise and enhances prediction accuracy.

\end{abstract}


\IEEEpeerreviewmaketitle

\section{Introduction}\label{sec:Intro}

\IEEEPARstart{T}{raffic} prediction is among the critical components of modern intelligent transportation systems (ITS).
It is an effective approach to analyzing future traffic trends based on urban road traffic conditions (e.g., speed, flow, and road interconnection relationships) \cite{TGCN,Yutraffic}.
An accurate and efficient traffic prediction system serves significant practical benefits, based on which traffic decisions empower a range of emerging applications, including but not limited to traffic resource allocation, traffic congestion prevention, and daily travel arrangements \cite{Urbancomputing, zhang2011data}. 
Therefore, providing reliable predictions of future traffic conditions is indispensable for building an advanced ITS.

In a typical setting, traffic data is a time-series recorded with fixed intervals by sensors deployed at specific locations \cite{ASTGNN}.
Intuitively, this type of data inherently holds two properties: 
1) Temporally correlated, where observed traffic data from the periods before and after are correlated and change over time. 
2) Spatially correlated, where traffic conditions on different roads are affected by adjacent ones due to the topological connectivity of urban traffic networks.
Aiming to provide accurate forecasting solutions, recent studies focus on the perspective of capturing both temporal and spatial dependence.

Regarding the former, a number of early works leveraged statistical and classical machine learning models to capture temporal correlations, such as linear-based methods (including auto-regressive integrated moving average \cite{ARIMA} and vector auto-regressive  \cite{VAR}) 
and machine learning-based methods (e.g., support vector regression \cite{SVR} and multi-layer perceptron \cite{MLP}).
The performance of these methods highly depends on the linearity assumption or hand-crafted data features and shows poor accuracy in practical scenarios \cite{zhang2011data}.
Considering the modeling capability of deep neural networks for time-series data dependence, researchers have designed a range of deep neural network-based methods for traffic prediction.
For example, convolutional neural network-based \cite{CNN_traffic}, recurrent neural network-based methods and their variants \cite{DCRNN}.

From the spatial dependence standpoint, a large volume of work adopted advanced graph convolutional network (GCN) to enhance the non-Euclidean data modeling capabilities, resulting in significant performance improvement in traffic forecasting tasks \cite{DL_traff}.
In a nutshell, these approaches extract the spatial dependence in traffic networks by building graphs and aggregating node embedding.
Studies like ASTGCN \cite{ASTGCN} first build a spatial topology graph based on the geographic distance between sensors and then adopt graph convolutional layers to learn spatial dependence. 
Subsequent efforts, such as GraphWaveNet \cite{GWNet}, dynamically create the spatial topology graph from node embedding and achieve significant prediction performance.
Due to the excellent ability to capture fine-grained spatial dependence, these graph-based methods are dominating traffic prediction tasks.

However, recent traffic prediction methods excessively focused on the spatial correlation of traffic data \cite{DL_traff}, resulting in two main drawbacks.
Firstly, graph-based approaches stack a number of spatial dependence capturing modules, leading to over-parameterization that makes the model complex and challenging to get well-trained.
Secondly, graph-based approaches rely on neural networks to automatically exploit temporal correlations from raw data, while they are inherently noisy and even incomplete \cite{impuation}.
It has also been shown in recent studies that deep neural networks tend to over-fit noisy data \cite{Universal,DL_traff}. 
Therefore, over-focusing on spatial dependence and modeling time-series indiscriminately may undermine the power of traffic prediction model.


Considering the aforementioned problems, we rethink the quality of the raw data fed into the traffic prediction model, aiming to enhance existing algorithms from the ground up. 
In this paper, we take advantage of filtering algorithms from the digital signal processing area, where the filters are designed to alleviate noise from the traffic data.
Our motivation stems from this assumption: time-series dependence may be more easily captured after filtering noise.
To verify this assumption, we first conduct a straightforward validation experiment in \cref{sec:analysis}.
The results show that the accuracy of the model can be significantly improved by a moving average filter.


Inspired by the empirical results, we further propose a universal learnable filter module.
The module converts the input to the frequency domain using Fast Fourier Transform (FFT) and then recovers the denoised data by inverse Fast Fourier Transform (IFFT).
In order to filter noise and extract meaningful features, we design the learnable filter based on parametric neural networks, which can automatically alleviate noisy features in the frequency domain.
To summarize, our contributions are as follows:
\begin{itemize}
    \item We propose an effective filter module to improve the traffic prediction performance, which converts the data to the frequency domain for denoising and then restores noise-free data.
    
    \item The proposed filter module is easy to implement and universal, which can be integrated into any existing state-of-the-art traffic prediction for free.
    
    \item We demonstrate the efficiency of the proposed filter module through extensive experiments on two real-world datasets. 
    Specifically, we conduct numerical experiments with five advanced prediction models, and the simulation results demonstrate that the proposed filter module can consistently enhance the prediction performance.
\end{itemize}

The remainder of this paper is organized as follows. 
\cref{sec:related} gives a brief overview of the related work.
\cref{sec:pre} formulates the traffic prediction problem. 
Then, we present an empirical analysis and describe the proposed filter module in \cref{sec:method}.
Subsequently, we conduct a series of case studies on two real-world traffic datasets and analyze the simulation results in \cref{sec:experiments}
Finally,  this work is concluded in \cref{sec:conclusion}.
\section{Related Work}\label{sec:related}
In recent years, various deep learning models have been proposed for traffic prediction and consistently outperform traditional approaches.
As traffic data is temporally and spatially coupled, these approaches improve prediction accuracy from two aspects, i.e., temporal dependence and spatial dependence.

Temporal dependence learning modules in traffic prediction approaches can be grouped into three categories, namely, recurrent units, convolutional layers, and attention mechanisms.  
Recurrent neural network (RNN) is proposed for learning temporal dependency in time-series and is further improved by its variants LSTM and GRU. 
LSTM and GRU are widely adopted in traffic prediction approaches, see DCRNN \cite{DCRNN} and DGCRN \cite{DGCRN} for examples.
Compared to RNN, temporal convolution network (TCN) efficiently exploit time-domain correlation information by utilizing convolutional filters\cite{STGCN, GWNet}. 
The attention mechanism, which originated in the field of natural language processing, has shown remarkable performance on sequence modeling tasks.
Some traffic prediction approaches employ this attention mechanism as an enhancement of conventional modules. 
For example, ASTGCN combines CNN layers with attention mechanism \cite{ASTGCN}, GMAN \cite{GMAN} learns temporal dependence with node embeddings and attention mechanisms. 
In addition, other researchers adopt self-attention mechanism and Transformer architecture to extract global dependence in the sequence, demonstrating advantages of learning long-term temporal dependence \cite{ASTGNN}.

For extracting spatial dependence in traffic data, the graph neural network (GNN) is ideally suited due to its superiority in learning spatial features in non-Euclidean space.
Sophisticated GNN structures have been proposed for the traffic prediction task in the past few years.
Preliminary traffic prediction studies employ conventional GNN modules on the geographic graph, e.g., graph convolution layers \cite{Yutraffic, ASTGCN} and graph attention networks \cite{ZhangGAT}. 
Recent studies achieve state-of-the-art performance by designing tailor-made graph construction methods, as it is not sufficient to only use a geographic graph to represent spatial dependence in traffic data.
Compared to preliminary approaches, GraphWaveNet \cite{GWNet} improves prediction performance by utilizing an adaptive adjacent matrix that does not rely on any prior geo-information.
MTGNN \cite{MTGNN} represents spatial dependence with an asymmetric graph, which is trained with learnable node embeddings.
Other studies extract spatial dependence by constructing heterogeneous graphs \cite{ASTGNN, STFGNN, DGCRN}.
For instance, STFGNN \cite{STFGNN} generates a temporal graph using dynamic time warping and merges the temporal graph with a geographic graph. 
DGCRN \cite{DGCRN} integrates a pre-defined static graph with a dynamic graph generated by filtering the node embedding to capture spatial dependencies.

To summarize, the aforementioned studies have long-term progress in traffic prediction via designing spatial-temporal neural network modules. Nevertheless, existing studies ignored the noise in traffic data, which may corrupt model training and undermine prediction accuracy.
Therefore, we propose a learnable filter module that adaptively filters out short-term fluctuations in time series.

\section{Preliminaries}\label{sec:pre}

In a traffic prediction problem, the observed traffic data (e.g., speed, flow) with $N$ sensors are denoted as $\mathbf{X} =\left\{\mathbf{X}^1,\mathbf{X}^2,\ldots,\mathbf{X}^T \right\}$ ($\mathbf{X} \in \mathbb{R}^{N\times T\times F}$), where $F$ is the feature dimension and $\mathbf{X}^t$ represents the observed value at the time step $t$.
The traffic forecasting problem involves predicting the state of future variables by using historical recorded values, i.e., learning a mapping function $f(\mathbf{X};~\theta)$ parameterized by $\theta$ to predict the future traffic state:
\begin{equation}\label{equ:predict}
   \left[\mathbf{X}^{t-H+1}, \mathbf{X}^{t-H+2}, \cdots, \mathbf{X}^t\right] \stackrel{f}{\longrightarrow}\left[\hat{\mathbf{X}}^{t+1}, \hat{\mathbf{X}}^{t+2}, \cdots, \hat{\mathbf{X}}^{t+T}\right],
\end{equation}
where $H$ represents the historical time steps and $T$ is the predicted time steps.

Particularly, for graph-based traffic prediction, we define the traffic network as $G=(\mathcal{V}, \mathcal{E}, \mathcal{A})$ where $\mathcal{V}$ is the set of nodes ($|\mathcal{V}|=N$), and each node can be considered the sensor in the traffic network;
$\mathcal{E}$ is the set of edges that presents the connectivity among $\mathcal{V}$;
Accordingly, the adjacency matrix of $G$ is denoted as $\mathcal{A} \in \mathbb{R}^{N \times N}$.
Therefore, we can rewrite the mapping function as $f(G, \mathbf{X};~\theta)$. 
Also, the objective of traffic prediction modeling is to minimize the following function:
\begin{equation}
		\arg \mathop {\min }\limits_{\theta}{L(\theta)} = \frac{1}{T}\sum  \limits_{i=1}^{T}\ell (\mathbf{X}^i,~\hat{\mathbf{X}}^i),
\end{equation}
where ${\ell}(\cdot,\cdot)$ denotes the loss function.
Following the common practice \cite{GMAN, GWNet}, we use the mean absolute error as the loss function, i.e., ${\ell}(x,\hat{x})=|x-\hat{x}|$.

\section{Empirical Analysis and Methodology}\label{sec:method}

In this section, we first present an empirical analysis of the traffic data to justify our motivation for the proposed module. 
Then we introduce the principle of the Fourier transform and the fast Fourier transform, based on which we elaborate the learnable filter module in detail.
In addition, we provide an explanation of the function of the proposed module from the perspective of periodic convolution.

\subsection{Empirical Analysis}\label{sec:analysis}

\begin{figure}[t]
  \centering
  \includegraphics[width=0.85\linewidth]{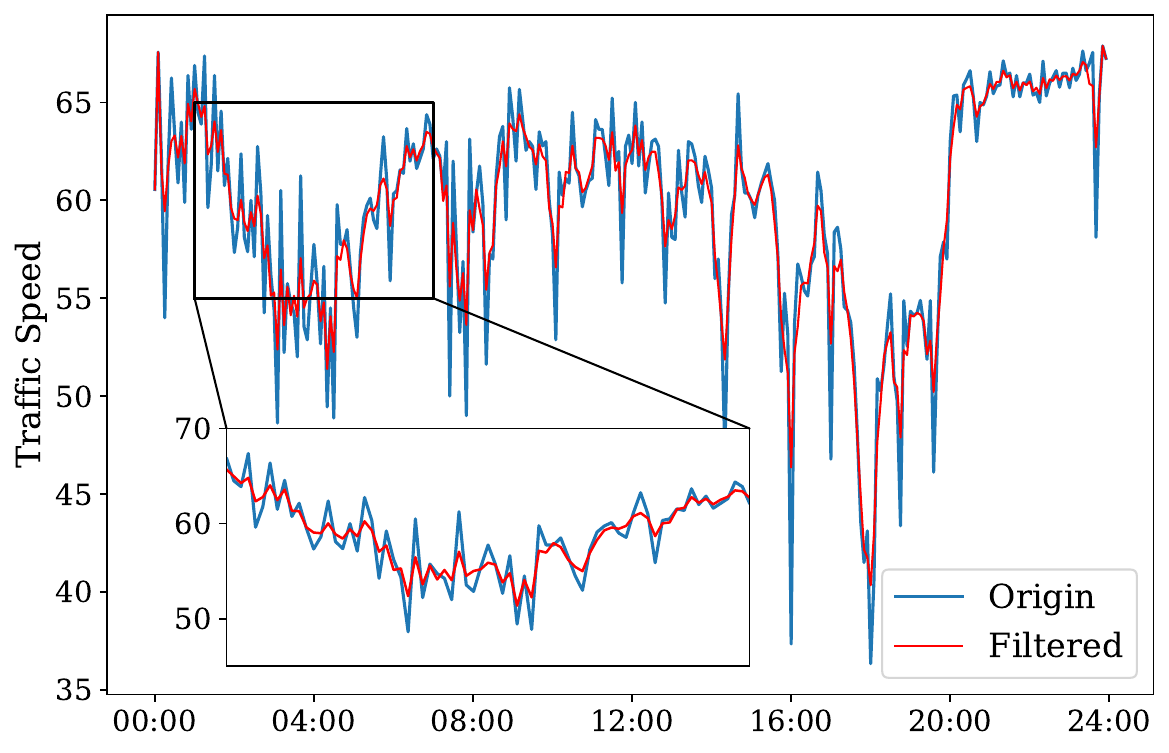}
  \caption{The traffic speed data was recorded at fixed nodes throughout the day. There are many sharp changes, which can be mitigated by a moving average filter.}
  \label{fig:datavisual}
\end{figure}

As previously stated, traffic data is recorded by fixed sensors, and such devices inherit inevitable noise from the environment when collecting data.
In \cref{fig:datavisual}, we show the data recorded by a particular sensor over the day, where it can be seen that the data change abruptly in many periods (presenting as sudden peaks and falls).
These unexpected changes may be caused by several factors, such as traffic accidents, sudden weather changes, or the breakdown of the sensor.
From a time-series perspective, this undesired variation can be considered as noise and is hard to track. As it is irregular and difficult to model due to data limitations.
In addition, a number of uncertainty quantification works have shown that predictive models have notably higher uncertainty where data change abruptly compared to other periods \cite{Uncertain,Bayes_traffic}.

Therefore, it is reasonable to assert that when predicting future traffic conditions, we should pay more attention to long-term trends over sudden changes (noises).
To further validate this idea, we conduct an empirical study on the METR-LA and PEMS-BAY datasets (specifications are presented in \cref{sec:exp_etup}) by a moving average filter.
Here, the moving average filter is a straightforward filtering algorithm that regards the successive sampled $N$ values as a queue and updates the original values with their averages.
By ``moving'' the sampling queue along the time axis (discarding values at the head of the queue and adding new data to the tail), the entire time-series can be mean-filtered.
In practice, we process the data with a moving average filter with a sample length of $5$ for noise filtering.
Then, take the mean of the filtered data and the original data to ensure the original trend and filter out the peaks.

The effect of the moving average filter on the traffic data is shown in \cref{fig:datavisual}, whose goal is to alleviate short-term fluctuations and highlight long-term trends.
To fairly validate the effect of filtering on the traffic data, we directly use the previous time step value to predict the next time step \cite{DL_traff}.
According to this method, the prediction in \cref{equ:predict} can be written as:
\begin{equation}
    \left[\hat{\mathbf{X}}^{t+1}, \hat{\mathbf{X}}^{t+2}, \cdots, \hat{\mathbf{X}}^{t+T}\right] \xrightarrow{f} \left[\mathbf{X}^{t}, \mathbf{X}^{t+1}, \cdots, \mathbf{X}^{t+T-1}\right].
\end{equation}
We choose this method to make predictions for the reason that it does not have any additional data manipulation, with the only control variable of the filter.

In this experiment, we use a filter with a sample length of $5$, filtering the past $1$ hour of data to predict the next hour.
We summarize the results in \cref{tab:exp_study}, from which we can see the accuracy improved after filtering.
This result indicates that the original data may contain noise that affects the performance of the prediction model, and using the filtering algorithm can relieve this problem.
In addition, the above results validate the assumption we made at \cref{sec:Intro}, i.e., time-series dependence may be more easily captured after filtering noise.

\begin{table}
    \caption{Improvement of Traffic Prediction Accuracy by Directly Using a Moving Average Filter}
    \centering
    \begin{tabular}{lcccccc}
    \toprule
    \multirow{2}{*}{Models}  &  \multicolumn{3}{c}{ METR-LA}  &
    \multicolumn{3}{c}{PEMS-BAY} \\
    \cmidrule(lr){2-4}    \cmidrule(lr){5-7}    
    & MAE & RMSE & MAPE & MAE & RMSE & MAPE\\
    \cmidrule(lr){1-7}
        Baseline  & 6.79 & 14.73 & 16.71 & 3.05 & 7.03  & 6.84  \\
        Filter  & \textbf{6.30} & \textbf{13.18} & \textbf{15.62}  & \textbf{2.90} & \textbf{6.79}  & \textbf{6.59}  \\
        \bottomrule
    \end{tabular}
    \label{tab:exp_study}
\end{table}

Inspired by the above results, we believe that the performance of models can be improved by filtering the raw input data.
However, an appropriate filter usually requires expert knowledge and extensive statistical experiments to identify the noise from the data.
Therefore, we propose an adaptive learnable filter, which transforms the data into the frequency domain, removes the noise in the frequency domain, and then recovers back to the time domain.
Before delving into the details of the learnable filter, we first introduce the basics knowledge of the Fourier transform.

\subsection{Fourier Transform}\label{sec:FFT}

\subsubsection{\textbf{Discrete Fourier Transform}}

Discrete Fourier transform is regarded as a fundamental technique in the field of digital signal processing, and is widely used in time-series data compression and spectrum analysis \cite{zhusemi}.
For the one-dimensional time-series DFT, denoted as $\mathbf{X} = \mathcal{F}\{\mathbf{x}\}$, its core idea is to convert a series of discrete digital sequences $\{\mathbf{x}_n\} :=\{x_0,x_1,\ldots,x_{N-1}\}$ into the frequency domain signal $\{\mathbf{X}_k\} :=\{X_0,X_1,\ldots,X_{N-1}\}$ by
\begin{subequations}\label{equ:DFT}
    \begin{align}
    X_k &=\sum_{n=0}^{N-1} x_n \cdot e^{-\frac{i 2 \pi}{N}  n k} \label{equ:DFT1} \\
    &=\sum_{n=0}^{N-1} x_n \cdot\left[\cos \left(\frac{2 \pi}{N}  n k\right)-i \cdot \sin \left(\frac{2 \pi}{N} n k\right)\right] \label{equ:DFT2},
    \end{align}
\end{subequations}
where $e$ is the natural constant and $i$ is the imaginary unit. 
\cref{equ:DFT} implies that DFT converts the time domain signals into a superposition of sine signals with different frequencies.
$X_k$ presents the spectrum of the sequence $\{\mathbf{x}_n\}$ at the frequency $\omega_k = 2 \pi k /N$.
For a given spectrum $X_k$, we can recover the original time domain signal by the inverse Discrete Fourier Transform (IDFT) as follows:
\begin{equation}\label{equ:IDFT}
    x_n =\frac{1}{N}\sum_{k=0}^{N-1} X_k \cdot e^{-\frac{i 2 \pi}{N} n k},
\end{equation}
which in practice means that $\{\mathbf{x}_n\}$ is expressed as a sum of different frequency components with $X_k$ as coefficients.

\subsubsection{\textbf{Fast Fourier Transform}}
Fast Fourier Transform (FFT) is an efficient and fast algorithm for computing DFT or IDFT of a sequence, which can significantly reduce the computational time complexity.
According to \cref{equ:DFT} and \cref{equ:IDFT}, we can observe that the time complexity required to compute DFT or IDFT is $\mathcal{O}(N^2)$.
FFT takes advantage of the symmetric and periodic nature of DFT and IDFT, to recursively decompose the computation of a length $N$ sequence into two shorter sequences $N_1, N_2$ ($N_1+N_2=N$). Thus, the time complexity is reduced to $\mathcal{O}(N\log N)$ \cite{cooley1965algorithm,VanFFT}.
Since FFT and DFT share the same principle of converting the input signal to the frequency domain where the periodic features are more easily distinguishable, it is commonly applied in digital signal processing to filter noisy signals in the time domain \cite{StaelinFFT,ZhouFilter}.
In this paper, we utilize FFT and filtering algorithms to alleviate the effect of noise in traffic data and then use inverse FFT (IFFT) to recover the data.

\subsection{Learnable Filter Module}\label{sec:Details_Filter}

\begin{figure}[t]
  \centering
  \includegraphics[width=0.99\linewidth]{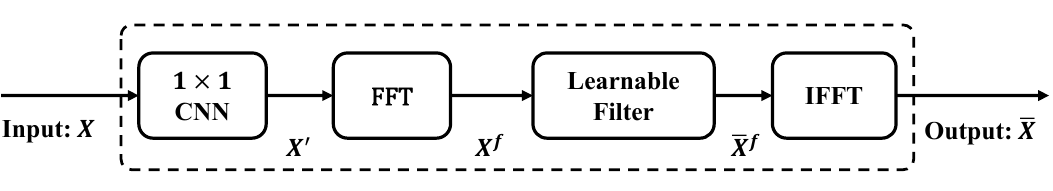}
  \caption{The structure of our proposed filter module, which consist of a $1
   \times 1$ CNN, a learnable filter kernel, and FFT, IFFT layers.}
  \label{fig:FilterModule}
\end{figure}

From \cref{equ:DFT}, we can find that $X_k$ is the frequency domain representation of the original time domain data. 
For time-series, signals may have different frequency components. 
Therefore, we can alleviate the noise-related frequencies in the frequency domain.
In other words, if we can design a network for learning and identifying the noise frequency, we can achieve the adaptive filter.
Based on this idea, we proposed a learnable filter module (shown in \cref{fig:FilterModule}), which includes a $1\times1$ CNN layer, FFT layer, learnable layer, and IFFT layer.
Here, the CNN functions to increase the non-linearity of the network, allowing the subsequent learnable filters to express more fine-grained features, and the FFT and IFFT serve to transform between the time and frequency domains.


For a given input $\mathbf{X} \in \mathbb{R}^{n \times d}$ ($n$ is the length of the sequence and $d$ represents the dimensions), we transform the time-series to the frequency domain by the FFT algorithm introduced in \cref{sec:FFT}:
\begin{equation}
    \mathbf{X}^f = \mathcal{F}(\mathbf{X}),
\end{equation}
where $\mathbf{X}^f \in \mathbb{C}^{n \times d}$ is a complex variable denoting the frequency domain representation (spectrum) of $\mathbf{X}$, $\mathcal{F}$ is FFT along the time axis.
After obtaining the frequency domain representation of the input, we can design a learnable filter for filtering the noise-related frequencies.
Specifically, we define a complex tensor $\mathbf{K}$, where $\mathbf{K} \in \mathbb{C}^{n \times d}$ consists of trainable parameters.
Then we element-wise multiply $\mathbf{X}^f$ and $\mathbf{K}$ to calculate :
\begin{equation}
    \overline{\mathbf{X}}^f =\mathbf{K} \odot \mathbf{X}^f.
\end{equation}
During the learning process, the parameters of  $\mathbf{K}$ can be iteratively updated by an optimization algorithm (e.g., stochastic gradient descent) to adaptively represent the noise filter in the frequency domain.
Finally, we recover the filtered spectrum $\overline{\mathbf{X}}^f$ into the time domain by IFFT and obtain the new time-series as follows:
\begin{equation}
    \overline{\mathbf{X}} =\mathcal{F}^{-1}(\overline{\mathbf{X}}^f),
\end{equation}
where $\overline{\mathbf{X}} \in \mathbb{R}^{n \times d}$, and $\mathcal{F}^{-1}$ is IFFT.
Through the above operations, the noise of the input data can be filtered, enabling the following prediction model to obtain a more accurate temporal representation. 

According to the convolutional theorem \cite{soliman1990continuous}, multiplication in the frequency domain is equivalent to the convolution of the time domain.
In particular, if two sequences have the same period, their operations can be considered as a circular convolution:
\begin{equation}\label{equ:circonv}
    (x * k)_{[n]} =\sum_{i=0}^{N-1}x_{[i]} \cdot k_{[n-i]},
\end{equation}
where $x$ and $f$ can be considered as the input and filter in the time domain, respectively.
Operator $*$ denotes the convolution in the time domain.
As a result, compared to the convolution in traditional neural networks, typically with a convolution kernel size of $3$ or $5$, the circular convolution has a larger receptive field over the whole sequence.
This design can better capture the whole sequence patterns and then filter the noise.

The motivation of literature \cite{ZhouFilter} is to simplify the over-parameterized architecture, which essentially serves as a replacement for the self-attention mechanism in the transformer structure, thus increasing its robustness.
Nevertheless, the design motivation of this paper is to filter noise from the temporal correlation perspective, which can be applied directly to the raw input data as a pre-processing module (this module can also be used as a substitution for convolution on the temporal dimension).
In other words, the noise in \cite{ZhouFilter} is the abnormal or malicious behavior of users, and noise in this paper is random variables that accompanies the data generation.

\section{Experiments}\label{sec:experiments}

\begin{table}
    \caption{Summary Statistics of METR-LA, PEMS-BAY, and PEMS08.}
    \centering
    \begin{tabular}{lccc}
    \toprule
    Dataset & \#Nodes  & \#Edges & \#Time steps\\
    \cmidrule(lr){1-4}
        METR-LA  & 207 & 1515 & 34272 \\
        PEMS-BAY & 325 & 2369 & 52166\\
        \bottomrule
    \end{tabular}
    \label{tab:Dataset}
\end{table}

\begin{table*}
\caption{Result Comparison of PEMS-BAY and META-LA}
\centering
\begin{tabular}{lccccccccc}
\toprule[1pt]
\multicolumn{10}{c}{METR-LA}\\
\cmidrule(lr){1-10}
\multirow{2}{*}[\multirowoffset]{Model} & \multicolumn{3}{c}{15 Minutes } & \multicolumn{3}{c}{ 30 Minutes } & \multicolumn{3}{c}{ 60 Minutes } \\ \cmidrule(lr){2-4}  \cmidrule(lr){5-7}  \cmidrule(lr){8-10} 
& MAE  & RMSE  & MAPE   & MAE  & RMSE  & MAPE   & MAE  & RMSE  & MAPE   \\ 
\cmidrule(lr){1-10}
CopyLastStep & 6.79  & 14.73  & 16.71  & 6.79  & 14.73  & 16.71 & 6.79  & 14.73  & 16.71 \\
Filter Module & 4.38  & 8.08  & 15.17  & 4.77  &  8.81  & 16.31  & 5.61  &  10.24  & 18.74 \\
\cmidrule(lr){1-10}
LSTNet    &  3.90 & 8.12   & 9.17 & 5.11  & 10.21 & 11.79   & 6.49  & 11.87 &  15.43          \\ 
LSTNet(+) & \textbf{3.70} & \textbf{8.04}  & \textbf{8.98}   & \textbf{4.62} & \textbf{10.13} & \textbf{11.43}  & \textbf{4.62}  & \textbf{10.13} & \textbf{11.43}       \\
\cmidrule(lr){1-10}
STGCN    & 3.47 & 7.99 & 8.57  & 4.26  & 9.95 & 10.70  & 5.08  & 11.80  & 13.12  \\ 
STGCN(+) & \textbf{3.36} & \textbf{7.85} & \textbf{8.33}  & \textbf{4.13}  & \textbf{9.87} & \textbf{10.34}  & \textbf{5.06}  & \textbf{11.65}  & \textbf{12.70}   \\
\cmidrule(lr){1-10}                       
ASTGCN    &   3.58 & 7.90 & 9.05    &  4.44 & 9.88 & 11.47   & 5.54 & 11.70 & 14.76  \\
ASTGCN(+) &   \textbf{3.39} & \textbf{7.62} & \textbf{8.66}    &  \textbf{4.09} & \textbf{9.48} & \textbf{10.55}   & \textbf{5.05} & \textbf{11.36} & \textbf{13.12}   \\
\cmidrule(lr){1-10}
GMAN      &   4.22 & 9.57 & 10.54   &  4.61 & 10.65 & 11.39   & 5.67 & 12.79 & 13.85 \\
GMAN(+)   &   \textbf{3.96} & \textbf{8.90} & \textbf{10.40}   &  \textbf{4.24} & \textbf{9.80}  & \textbf{10.99}   & \textbf{5.15} & \textbf{11.76} & \textbf{13.14}   \\
\cmidrule(lr){1-10}
\cmidrule(lr){1-10}
GraphWaveNet   & 3.23 & 7.56 & 7.80   &  3.97 & 9.55 & 9.82  &  4.92 & 11.64 & 12.47  \\
GraphWaveNet(+)   & \textbf{3.18} &\textbf{7.32} & \textbf{7.78}   &  \textbf{3.90} & \textbf{9.34} & \textbf{9.71} & \textbf{4.85} & \textbf{11.56} & \textbf{12.15} \\
\bottomrule
\toprule
\multicolumn{10}{c}{PEMS-BAY}\\ 
\cmidrule(lr){1-10}
\multirow{2}{*}[\multirowoffset]{Model} & \multicolumn{3}{c}{15 Minutes } & \multicolumn{3}{c}{ 30 Minutes } & \multicolumn{3}{c}{ 60 Minutes } \\ 
\cmidrule(lr){2-4}  \cmidrule(lr){5-7}  \cmidrule(lr){8-10} 
& MAE  & RMSE  & MAPE   & MAE  & RMSE  & MAPE   & MAE  & RMSE  & MAPE   \\ 
\cmidrule(lr){1-10}
CopyLastStep & 3.05  & 7.03  & 6.84  & 3.05  & 7.03  & 6.84 & 3.05  & 7.03  & 6.84 \\
Filter Module  & 2.22  & 5.01  & 5.50  &  2.69  & 6.05  & 6.60  & 3.36 & 7.49  & 8.24 \\
\cmidrule(lr){1-10}
LSTNet    &  1.63 & 3.24  & 3.40  & 2.45  & 4.43 & 5.24   & 2.87  & 5.26 & 6.43          \\ 
LSTNet(+) &  \textbf{1.49} & \textbf{3.07}  & \textbf{3.20}  & \textbf{1.89}  & \textbf{3.96} & \textbf{4.33}   & \textbf{2.25}  & \textbf{4.64} &  \textbf{5.42}          \\ 
\cmidrule(lr){1-10}
STGCN    &  1.33 &  2.83  & 2.80  & 1.70  & 3.89 & 3.89   & 2.06  & 4.75 & 5.02          \\ 
STGCN(+)    &  \textbf{1.33} & \textbf{2.83}  & \textbf{2.79}  & \textbf{1.70}  & \textbf{3.81} & \textbf{3.83}   & \textbf{2.04}  & \textbf{4.69} & \textbf{4.91}         \\ 
\cmidrule(lr){1-10}                       
ASTGCN   & 1.44 & 3.06 & 3.25   &  1.80 & 4.07 & 4.40  &  2.10 & 4.77 & 5.30   \\
ASTGCN(+) & \textbf{1.43} & \textbf{3.06} & \textbf{3.22}  &  \textbf{1.75} & \textbf{3.96} & \textbf{4.18}  & \textbf{2.03} & \textbf{4.62} & \textbf{4.95}  \\
\cmidrule(lr){1-10}
GMAN  & 1.80 & 4.22 & 4.47   &  1.80 & 4.14 & 4.42  &  2.19 &  5.03 & 5.29    \\
GMAN(+) & \textbf{1.67} & \textbf{3.82} & \textbf{3.99}    &  \textbf{1.72}  & \textbf{3.93}  & \textbf{4.09}  &  \textbf{2.19}  &  \textbf{5.01}  & \textbf{5.23} \\
\cmidrule(lr){1-10}
\cmidrule(lr){1-10}
GraphWaveNet    &  1.32 & 2.76 & 2.78 & 1.66 & 3.74 & 3.75 & 1.99 & 4.56 & 4.75   \\
GraphWaveNet(+)   & \textbf{1.31}  &  \textbf{2.74}  &   \textbf{2.77} & \textbf{1.65} & \textbf{3.70} & \textbf{3.74} & \textbf{1.95} & \textbf{4.39} & \textbf{4.60} \\
\bottomrule[1pt]
\end{tabular}
\label{tab:overall}
\end{table*}

In this section, we evaluate the proposed filter module by performing comprehensive case studies on two real-world traffic datasets. 
Specifically, we first present the preliminary content, such as the dataset, the evaluation metrics, and the baseline algorithms.
Then we demonstrate the performance improvement of the proposed module for all investigated algorithms.
Next, we take GraphWaveNet as a representation to verify the capability of the filter module and compare the time consumption.
Finally, we also analyze and visualize the result of the proposed module.

\subsection{Setup}\label{sec:exp_etup}

\subsubsection{Dataset}
We verify the performance of the proposed filter module on two public traffic network datasets, namely, the METR-LA \cite{DCRNN} and PEMS-BAY.
METR-LA is a traffic dataset for Los Angeles County, containing speed information collected from sensors on $207$ freeways from 3/1/2012 to 6/30/2012.
PEMS-BAY is a traffic flow dataset collected from California Transportation Agencies Performance Measurement System (PEMS). 
It covers $6$ months (from 1/1/2017 to 5/31/2017) of data collected in the Bay Area with $325$ traffic sensors.
All data are collected at \SI{5}{\minute} intervals, we provide detailed statistics of the dataset in \cref{tab:Dataset}.

\subsubsection{Baseline Methods}
We select a number of representative advanced traffic prediction algorithms to verify the effectiveness of the proposed module, including:
\begin{itemize}
    \item \textbf{LSTNet}: LSTNet is an auto-regressive neural network model based on CNN and RNN components \cite{LSTNet}.
    
    \item \textbf{STGCN}: STGCN is a spatio-temporal graph neural network with GCN to capture spatial features and CNN to capture temporal dependencies \cite{STGCN}.
    
    \item \textbf{ASTGCN}: ASTGCN is an attention-based spatio-temporal graph convolutional networks, which design spatial attention and temporal attention mechanisms for spatio-temporal modeling \cite{ASTGCN}.
    
    \item \textbf{GMAN}: GMAN is an Encoder-decoder model consisting of graph-based multi-headed attention networks \cite{GMAN}.
    
    
    \item \textbf{GraphWaveNet}: GraphWaveNet is designed to capture spatio-temporal dependencies by stacking temporal convolution and adaptive graph convolution modules \cite{GWNet}.
\end{itemize}
In addition, we also show the performance of only using the learnable filter module as the prediction model, which can be compared with the algorithm (\textbf{CopyLastStep}) we apply in \cref{sec:analysis}.

The implementation and data processing of all baseline algorithms in this paper follow the publicly available code\footnote{https://github.com/deepkashiwa20/DL-Traff-Graph} provided in the literature \cite{DL_traff}.
In practice, the module proposed in this paper is integrated as a pre-module before the baseline algorithm.
We conduct all experiments on a server with NVIDIA GeForce RTX 2080Ti GPUs.
For a fair comparison, we adopt the mean of $5$ runs of the algorithm as the presented report results.

\subsubsection{Metrics}
With respect to accuracy comparison, we evaluate the prediction accuracy of all methods using mean absolute error (MAE), root mean square error (RMSE), and mean absolute percentage error (MAPE) as metrics. 
These three metrics are widely used in traffic forecasting studies \cite{TGCN,ASTGCN,GWNet}, and they are defined as follows:

\begin{equation}
\begin{aligned}\
     {\rm MAE}(\mathbf{X},~\hat{\mathbf{X}})&=\frac{1}{N}\sum_{i}^{N}\left|{X^{(i)}-\hat{X}^{(i)}}\right|,\\
     {\rm RMSE}(\mathbf{X},~\hat{\mathbf{X}})&=\sqrt{\frac{1}{N}\sum_{i}^{N}\left(X^{(i)}-\hat{X}^{(i)}\right)^{2}},\\
    {\rm MAPE}(\mathbf{X},~\hat{\mathbf{X}})&=\frac{1}{N} \sum_{i}^{N}\left|\frac{X^{(i)}-\hat{X}^{(i)}}{X^{(i)}}\right| \times 100\%,
\end{aligned}
\end{equation}
where $X^{(i)}$ and $\hat{X}^{(i)}$ are the observed and predicted traffic value at time $i$, respectively.

\subsection{Overall Performance}
We first examine the performance of the baseline method and its combination with the proposed learnable filter.
Specifically, we compare their multi-step prediction performance on the METR-LA and PEMS-BAY datasets for future 15-, 30-, and 60-minute traffic conditions.
Among them, we use the `(+)' symbol to denote a variant of the baseline model equipped with the learnable filter.
For example, LSTNet(+) denotes the combination of LSTNet and the proposed filter. 
It should be noted that the base model of LSTNet(+) is identical to LSTNet, with the only difference being that a filter module is added before the model input.
\cref{tab:overall} summarizes the experimental results for all selected models, from which we can draw the following insights:
\begin{itemize}
    \item With the incorporation of the filter module, all models show a different level of improvement. 
    This is because the proposed filter module can alleviate the noise of the original data, enhancing the capability of the base model for capturing temporal dependence.
    The improvement scale on different datasets can further validate this viewpoint.
    Among them, the PEMS-BAY dataset holds high data quality and less noise \cite{DL_traff}, resulting in the effect of the filter is not obvious.
    In contrast, the METR-LA dataset inherent higher noise and outliers, enabling the filter to demonstrate its effectiveness significantly.
    
    \item For advanced GCN-based prediction models, the filtering modules of ASTGCN and GMAN show more remarkable improvements than STGCN and GraphWaveNet.
    The reason is that the former ones mainly concern with exploiting spatial dependencies, while the latter ones focus more on fine-grained temporal feature capturing.
    Therefore, we can directly integrate the proposed module with a GCN-based model to improve the capability for temporal dependence modeling.
    In addition, we conduct a representation case study on GraphwaveNet in \cref{sec:specialstudy} to analyze the capabilities of the filtering module and the underlying reasons for its less appreciable performance.
    
    \item Owing to the proposed filter module can filter out unnecessary noise and fit future traffic conditions with a more appropriate trend.
    When we adopt an individual filter module as the prediction model, it presents a significant improvement compared to CopyLastStep without additional operations.
    Meanwhile, a standalone filter module fails for long-term (60 minutes) predictions, which means that more advanced models need to be integrated.
    Certainly, we visualize and analyze the processing results of the proposed module in \cref{sec:vis}.
\end{itemize}

\subsection{Representative Case Study of GraphWaveNet}\label{sec:specialstudy}
\begin{figure}[t]
  \centering
  \includegraphics[width=0.8\linewidth]{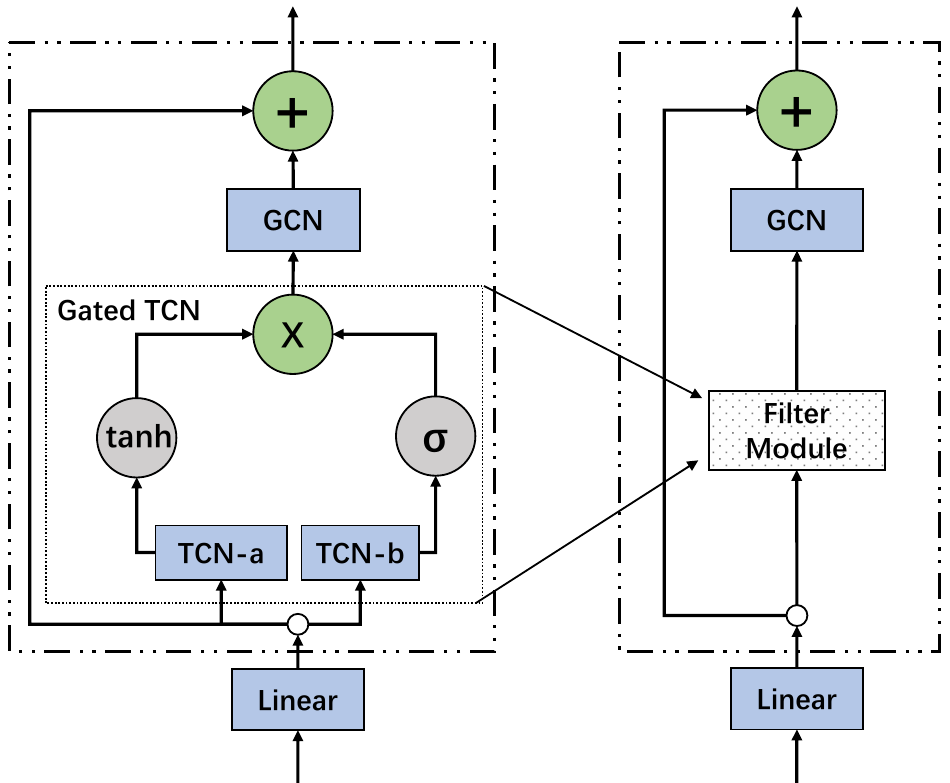}
  \caption{A GraphWaveNet structure consists of multiple stacks of the above modules. We replace the Gated TCN with the filter module.}
  \label{fig:Gwnet}
\end{figure}

According to the results shown in \cref{tab:overall}, the filter module is expected to enhance the performance advanced traffic prediction models, whereas there is no appreciable improvement when combined with the GraphWaveNet structure \cite{GWNet}.
Therefore, we further investigate the reasons why the filter module less boosting the performance of GraphWaveNet in this section.
As shown in the left of \cref{fig:Gwnet}, GraphWaveNet consists of stacked spatio-temporal blocks, each of which contains a GCN and a gated temporal convolutional layer (Gated TCN).
The idea of this architecture is to use Gated TCN to capture temporal dependencies and GCN to capture spatial dependencies \cite{GWNet}.

\begin{table}
\caption{Comparison the Performance of Replacing Gated TCN with the Filter Module.}
\centering
\label{tab:GWnet_comp}
\begin{tabular}{llccl}
 \toprule
Temporal module & MAE & RMSE & MAPE & Time/epoch\\
\cmidrule(lr){1-5}
Baseline (8 Gated TCNs)   & 3.94 & 9.53 & 9.74  & 1.65s (100\%)  \\
8 Filter Modules   & 3.89 & 9.34 & 9.65  & 1.47s (91\%)  \\
6 Filter Modules    & 3.93 & 9.45 & 9.71   & 1.24s (75\%)  \\
4 Filter Modules     & 3.97 & 9.54 & 9.83   & 0.94s (57\%)  \\
\bottomrule
\end{tabular}
\end{table}

\begin{figure}[t]
  \centering
  \includegraphics[width=0.8\linewidth]{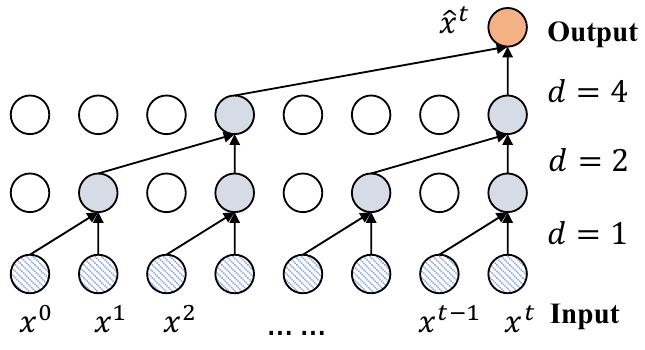}
  \caption{The 1-D dilated convolution with a kernel size of 2 \cite{GWNet}.}
  \label{fig:DilatedCNN}
\end{figure}

In this section, we examine the role of filters in traffic prediction tasks by replacing Gated TCN with the proposed filter module (see \cref{fig:Gwnet} for more details).
Besides, by replacing gated TCNs as filter modules, we further investigate the effect of stacked filter modules on performance and time consumption.
Specifically, we investigate the performance and the time required by decreasing the amount of time processing modules.
Please note that the typical GraphWaveNet employs $8$ time processing modules.

The experimental results are summarized in \cref{tab:GWnet_comp}, from which we can observe that replacing Gated TCN with the filter module can yield a modest improvement in the prediction accuracy.
Although this performance improvement is not impressive compared to other baselines, it still demonstrates that the learnable filter can be adopted as a replacement for the temporal processing module in GraphWaveNet.
The reason is that Gated TCN captures fine-grained temporal dependencies by integrating two parallel TCNs, which are implemented by expanding causal convolution.
\cref{fig:DilatedCNN} depicts a 1-D dilated causal convolution structure with a kernel size of $2$.
The dilated convolution integrates a dilation factor to the traditional convolution kernel (i.e., convolution with holes), which can widen the receptive field exponentially.
With stacked dilated causal convolution, GraphWaveNet can process temporal dimensional features with a larger horizon \cite{TCN}.
Based on the above analysis and experimental results, we can conclude that the design principle of the dilated convolution is consistent with our analysis in \cref{sec:Details_Filter}, where the learnable filter is a convolutional layer with a global receptive field.

Furthermore, we also find that replacing the filter module can reduce the parameters and computation time while ensuring comparable performance.
As presented in \cref{tab:GWnet_comp}, the time consumption of the model can be reduced by $9\%$ after replacing Gated TCN with the filter module. 
By further reducing filter modules, it can achieve effective performance with $57\%$ of the baseline time spent.
Therefore, replacing the temporal processing module in GraphWaveNet with filter modules can ensure the prediction performance, while greatly  reducing the number of parameters and improving the efficiency.

\begin{figure}[t]
  \centering
   \includegraphics[width=0.85\linewidth]{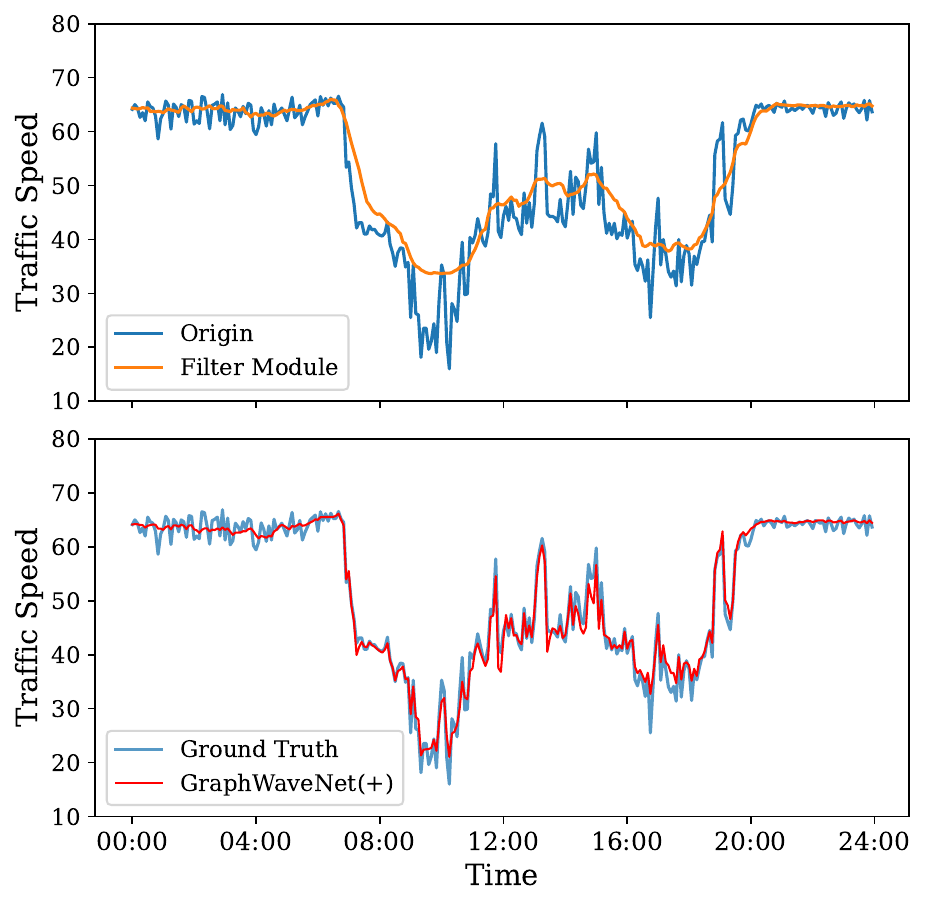}
  \caption{Processing results of the filter module (top); Prediction of GraphWaveNet with filter module integrated (bottom).}
  \label{fig:interpre}
\end{figure}

\subsection{Filter Module Interpretation}\label{sec:vis}
In this section, we analyze and visualize the result of the proposed filter module.
In \cref{fig:interpre}, we depict the speed record of a node in the METR-LA dataset over a day, showing both the filtered data and the predictions of GraphWaveNet(+).
Regarding the filter module, we can figure that the filter module is equivalent to a smoothing process of the original data, which ignores the sharp rises and falls in the short term.
This observation indicates that the proposed filter module holds the property of noise mitigation.
However, if we only adopt the filtered data as a prediction, the accuracy is not satisfactory.
The reason is that while the individual filter module alleviates noise and emphasizes long-term trends, it inevitably leads to bias.


Therefore, we believe the proposed filter module should be incorporated into advanced existing models to better capture fine-grained temporal dependencies while filtering out noises.
In \cref{fig:interpre}, we also present the prediction performance of the GraphWaveNet model integrating the filter module.
It can be easily observed that GraphWaveNet(+) can produce stable and accurate predictions.
For small microscopic fluctuations, the predicted values always follow the mean of the ground truth.
For larger fluctuations, the predicted values can fit accurately rather than just providing smoothed values.
The above results are attributed to the noise mitigation and the ability of GraphwaveNet to model long- and short-term temporal dependencies, 
which can fully demonstrate the effectiveness of integrating the filter module and existing models.

\section{Conclusion}\label{sec:conclusion}

In this paper, we propose a learnable adaptive filter to filter out short-term fluctuations in time-series for traffic prediction.
The proposed module first converts the data into the frequency domain by FFT.
In the frequency domain, a parametric neural network is designed to mitigate the noise-dependent frequencies.
Finally, we recover the denoised data to the time domain by IFFT.
We conduct comprehensive experiments on two public datasets, and the results demonstrate that the proposed module, as a universal pre-processing module, can be integrated into existing traffic prediction models and enhance prediction performance.
In addition, we make a thorough analysis of the filter module and provide an in-depth case study on GraphWaveNet. 
We show that the proposed module can not only filter the noise as well as serve as a temporal processing module in the spatio-temporal prediction model.
In the future, we will combine the filter and transformer models to build a GCN-free traffic prediction model.

\bibliographystyle{IEEEtran}
\bibliography{ref}

\ifCLASSOPTIONcaptionsoff
  \newpage
\fi

\end{document}